\documentclass[final,5p,times,twocolumn]{elsarticle}

\usepackage{amssymb}
\usepackage{amsmath}
\usepackage{hyperref}
\hypersetup{hidelinks=true}

\journal{Engineering Applications of Artificial Intelligence}
\begin{document}
\begin{frontmatter}

\title{ViscNet: Vision-Based In-line Viscometry for Fluid Mixing Process}
\author[1]{Jongwon Sohn\corref{equal}}
\author[2]{Juhyeon Moon\corref{equal}}
\author[1]{Hyunjoon Jung}
\author[1,3]{Jaewook Nam\corref{pi}}
\ead{jaewooknam@snu.ac.kr}
\cortext[equal]{Equal Contribution}
\cortext[pi]{Corresponding Author}
\affiliation[1]{organization={Seoul National University},
            addressline={Department of Chemical and Biological Engineering},
            city={Gwanak-gu},
            postcode={08826},
            state={Seoul},
            country={Korea}}
\affiliation[2]{organization={Seoul National University},
            addressline={Department of Mechanical Engineering},
            city={Gwanak-gu},
            postcode={08826},
            state={Seoul},
            country={Korea}}
\affiliation[3]{organization={Seoul National University},
            addressline={Institute of Chemical Process},
            city={Gwanak-gu},
            postcode={08826},
            state={Seoul},
            country={Korea}}

\begin{abstract}
Viscosity measurement is essential for process monitoring and autonomous laboratory operation, yet conventional viscometers remain invasive and require controlled laboratory environments that differ substantially from real process conditions. We present a computer-vision-based viscometer that infers viscosity by exploiting how a fixed background pattern becomes optically distorted as light refracts through the mixing-driven, continuously deforming free surface. Under diverse lighting conditions, the system achieves a mean absolute error of 0.113 in $\log\mathrm{m}^2 \mathrm{s}^{-1}$ units for regression and reaches up to 81\% accuracy in viscosity-class prediction. Although performance declines for classes with closely clustered viscosity values, a multi-pattern strategy improves robustness by providing enriched visual cues. To ensure sensor reliability, we incorporate uncertainty quantification, enabling viscosity predictions with confidence estimates. This stand-off viscometer offers a practical, automation-ready alternative to existing viscometry methods.
\end{abstract}









\end{frontmatter}


\section{Introduction}
\label{intro}
Viscosity is a fundamental property of fluids that quantifies the internal friction generated within the fluid flow. Because it strongly influences the characteristics of fluid motion, viscosity is measured across a broad range of laboratory and industrial settings in food processing~\citep{Mathijssen2023}, coating operations~\citep{Hawley2019}, pharmaceuticals manufacturing~\citep{Vilimi2024, Zidar2020}, and petrochemical production~\citep{Xu2022}. In automated systems, viscosity functions not only as a diagnostic indicator of process conditions~\citep{Ferraz2022, Su2022} but also as a key parameter guiding how fluids should be manipulated and controlled~\citep{Walker2023}. Therefore, the ability to measure viscosity rapidly, reliably, and with minimal operational burden has become increasingly critical.

Modern viscometry is dominated by mechanical instruments that induce relative motion between a sensor component and the fluid, infering viscosity by measuring the resulting frictional force, velocity change, or other dynamic response of the internal component. Classical examples include falling-ball, capillary, and spindle viscometers~\citep{Bhattad2023}. However, these instruments require highly idealized, laboratory-specific environments that differ substantially from real process conditions. Hence, fluid samples must be extracted and transported to the device, a procedure that can be time-consuming and may compromise the sterility or purity of the original content~\citep{Nour2020}. In addition, sampling errors may arise from spatial inhomogeneity within the fluid volume or from changes that occur during the sampling interval~\citep{Hilliou2020}.

To overcome these issues, in-line viscometers have been developed to be installed directly in pipelines and to measure viscosity during operations. While these instruments avoid the need for off-line sampling, they still require direct contact with the fluid, which presents several challenges. Sensor lifetime can be compromised by aggressive flow conditions, elevated temperature, or high viscosity~\citep{Steiner2010}, and the measurement inherently reflects only a partial local state rather than the true bulk property. In addition, installation, removal, and periodic recalibration often require significant effort, further limiting their practicality~\citep{Gorey2024}. These challenges highlight the need for stand-off viscometers that can infer viscosity without any direct interaction with the fluid.

A promising alternative is to use computer-vision techniques. Recent studies have explored the idea that viscosity is reflected in the motion of the free surface of the fluid. It has been shown from a biological perspective that the shape of a fluid surface contains visually perceptible cues about its viscosity~\citep{VanAssen2020}. Various vision-based viscometry methods building on this insight have emerged, including the use of robotic interaction to generate surface motion in a bottle and estimate viscosity~\citep{Park2024, Walker2023}, or using top-view videos of sloshing fluid in the cup to compute optical flow and infer viscosity~\citep{An2021}. A key challenge shared across these works is the need to effectively visualize the dynamics of the fluid surface. To obtain sufficiently clear observations, these methods rely on controlled or specialized experimental conditions. Consequently, although they offer stand-off measurements, they remain difficult to deploy in practical \textit{in-situ} settings.

In this work, we address the limitations of previous viscometers by extending computer-vision-based viscometry to fluid mixing processes. Mixing is a widely used operation in fluid-related processes~\citep{Barabash2018, Jaszczur2020}, and its free surface naturally exhibits rich fluctuation and decay behaviors that make it well suited for optical analysis~\citep{Haque2006}. These surface dynamics provide a physically meaningful and easily observable signal from which viscosity can be inferred without intrusive instrumentation. Motivated by these observations, we present ViscNet, a computer-vision-based, in-situ, stand-off viscometer designed specifically for mixing environments. The main contributions of this work are summarized as follows:

\begin{itemize}
    \item We demonstrate that bulk viscosity can be inferred by analyzing how viscous dissipation modulates the surface geometry of the stirring-induced flow.
    \item We show that a background pattern acts as an optical probe whose refraction-induced distortion through the fluid surface encodes viscosity-dependent dynamics.
    \item We constructed a large-scale video dataset of stirred fluids, combining real experimental recordings with simulation data generated via particle-based fluid simulations and photorealistic rendering.
    \item We developed a modified Video Vision Transformer (ViViT) architecture that predicts viscosity directly from videos of vortex surface with impeller rotation speed embedded as an additional conditioning signal.
\end{itemize}

The contents of the paper are organized as follows. Section 2 reviews previous work on stirring-flow physics and transparent free-surface visualization. Section 3 presents our methodology, including the model architecture, dataset construction, and training schemes. Section 4 reports viscosity estimation results, analyzes model behavior, and outlines additional experiments for improvement. Section 5 concludes with a summary of our findings, discusses the significance and limitations of our model, and highlights directions for future work.

\section{Related Works}
\label{sec2}
\subsection{Viscous Effects on Free Surface of Mixing Vortex}
\label{related:vortex}
The critical influence of viscosity on the shape and motion of stirring vortex surfaces has been demonstrated across theoretical models, experimental investigations, and simulation studies. This influence is particularly evident during impeller-driven mixing, where the internal flow field separates into an inner forced-vortex region governed directly by the impeller and an outer free-vortex region shaped primarily by the container walls~\citep{Haque2006}. In the outer region, wall friction generates small turbulent eddies, producing corrugated surfaces whose size and morphology are governed primarily by viscous forces. Also, mixing vortices exhibit polygonal or multi-lobed surface motions whose size, shape, and rotation rate vary with viscosity~\citep{Bach2014, Jansson2006, Tophoj2013}. Furthermore, when external forcing of rotation is removed, the dissipation of vortex rotation and free-surface curvature is likewise viscosity dependent~\citep{Maas1993, Siginer1984}. 

\subsection{Visualization of Transparent Free Surface}
\label{related:surface}
Visualizing transparent fluids has long been a challenging task due to their background-dependent appearance. Prior works primarily rely on segmentation-based approaches, using motion cues, appearance cues, or alternative imaging to isolate transparent liquids across various setups~\citep{Eppel2021, Liao2020, Schenck2017, Yamaguchi2016}. Our method takes more suitable and versatile approach by adopting the background-oriented schlieren (BOS) imaging principle, where the interface geometry of a transparent medium is inferred from optical distortion of a background pattern~\citep{Raffel2015}. Mathematical models and, more recently, neural network methods have further enabled dynamic free-surface reconstruction from BOS sequences~\citep{Raffel2023,Thapa2020}. Finally, reflection-based signals can also provide an additional source of surface information. Temporal variations in specular reflections have been shown to encode surface vibrations in a previous study~\citep{An2021}. In mixing flows where smooth (laminar) and highly corrugated (turbulent) regions coexist~\citep{Haque2006}, such reflections range from localized specular flashes to diffuse highlights outlining the interface contour, providing complementary cues to BOS distortions in characterizing vortex surface dynamics.

\section{Methods}
\label{sec3}
\subsection{Problem Definition}
\label{methods:definition}
We conceptualize the vortex generated during mechanical stirring as a time-varying hydrodynamic function $g$, driven by intrinsic fluid properties-viscosity $\nu$ and surface tension $\sigma$-and external agitation conditions such as impeller rotation speed $\omega$. The function $g$ produces the surface geometry $S$, which is then passed to an optical mapping function $f$, producing the visuals of the vortex surface. The background pattern $p$ also serves as an input to this operator. In this framework, the fluid interface $S$, whose geometry reflects the transfer of viscous momentum, acts as a refractive lens that distorts the input signal $p$. The distorted output is captured by a camera at time $t$ as an image $I_t$, and the sequence $\{I_t\}_{t=0}^{T}$ constitutes the video $V$. Thus, we solve an inverse problem: inferring the viscosity of an unknown fluid from the modulated output of the input signal.

In practice, the observed images are further perturbed by a nonparameterized noise source $\varepsilon$, arising from lighting fluctuations, surface bubbles, glass-induced distortions, and camera aberrations. These effects complicate the relationship between the physical mapping $f$ and its measured output.

\begin{equation}
\begin{aligned}
S \; &=\; g(\nu, \sigma;\,\omega) \;+\; \varepsilon_{\text{fluid}}, \\[6pt]
V \;=\; \{ I_t \}_{t=0}^{T} \; &=\; f(S;\, p) \;+\; \varepsilon_{\text{camera}}, \\[6pt]
\varepsilon_{\text{total}} \; &\sim\; \mathcal{N}\!\left(0,\; \sigma^{2}(x)\right).
\end{aligned}
\label{eq:problem_formulation}
\end{equation}

The primary objective of this study is to train a deep neural network ViscNet, parametrized by $\theta$, to invert the optical mapping $f$ and hydrodynamic mapping $g$: given the stirring condition $\omega$ and the video $V$, the model predicts the viscosity of the fluid as $\hat{\nu}$.

\begin{equation}
\begin{aligned}
\hat{\nu} &= \mathrm{ViscNet}_{\theta}[V;\omega]. \\[4pt]
\end{aligned}
\label{eq:problem formulation 2}
\end{equation}

\subsection{Video Vision Transformer Model}
\label{methods:model}

\begin{figure*}[t]
\centering
\includegraphics[scale = 0.9]{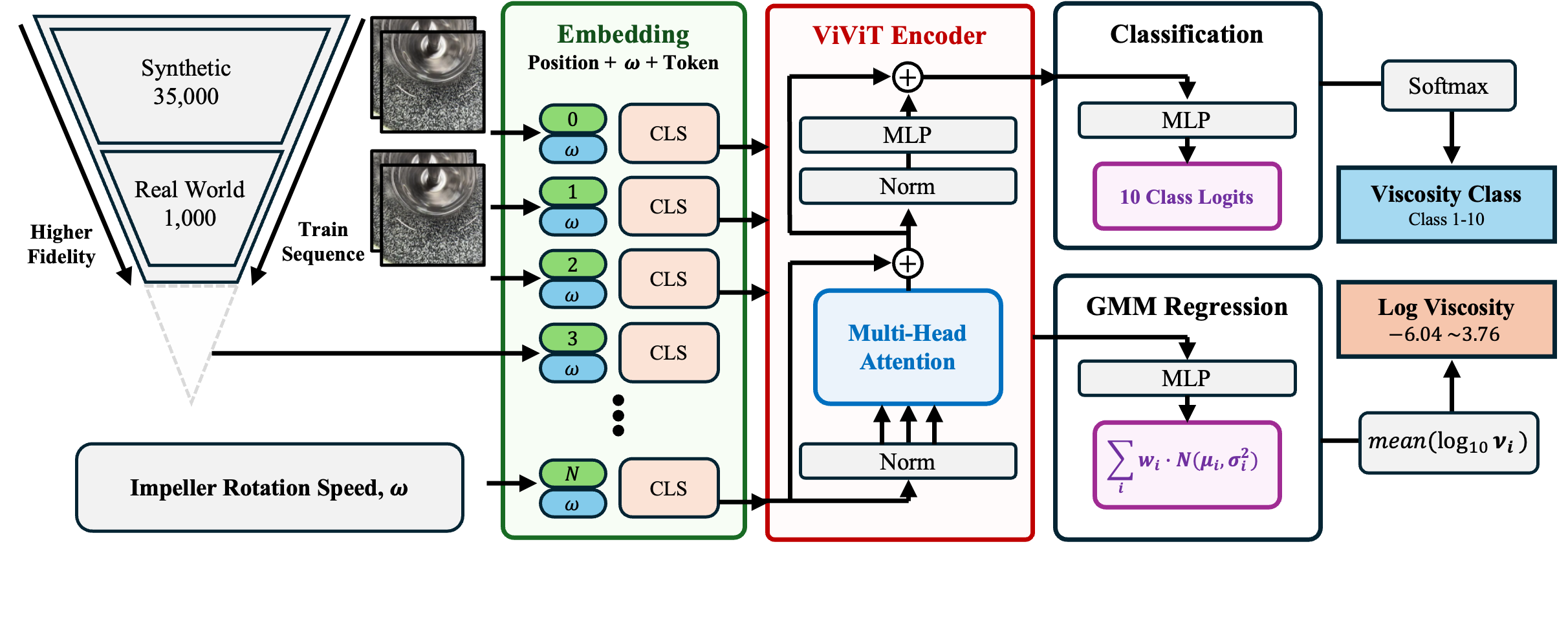}
\caption{ViscNet Model Architecture, Regression units are in log $\mathrm{m}^2 \mathrm{s}^{-1}$.}
\label{fig:modelstructure}
\end{figure*}

We present a viscosity estimator model, ViscNet, a deep neural network architecture based on Video Vision Transformer (ViViT; \citealp{Vivit2021}). ViViT is a state-of-the-art transformer model for video related tasks, developed by Google. A key addition to the original ViViT model is the incorporation of the impeller rotation speed $\omega$, which is embedded into the video input $V$ within the positional and token embedding layers. While impeller rotation drives the inertial flow, viscosity dictates the rate of momentum dissipation and the resulting spatial scales of the surface deformation. Embedding the rotation speed allows the model to decouple the input energy from the viscosity-dependent decay characteristics observed on the interface. It is also a quantity that is readily available in practical systems through close-loop motor control.

First, we implement a classification layer on top of ViViT to enable curriculum learning and to provide an interpretable training structure. We then replace this layer with a regression module based on a Gaussian mixture model (GMM) to quantify predictive uncertainty. During experiments, we observe that aleatoric heteroscedasticity contributes substantial error; therefore, we adopt a three-component GMM that explicitly reflects this behavior. Using three Gaussians supports multimodality while remaining compact enough to ensure stable training under negative log-likelihood (NLL) loss. The network deterministically outputs the mixture weights as well as the mean and standard deviation of each Gaussian component. This formulation is essential for the model to function as a sensor, as it allows us to assess reliability and evaluate empirical coverage. The general structure of the model is illustrated in Fig.~\ref{fig:modelstructure}.

\begin{equation}
\begin{aligned}
p(\nu \mid V,\omega)
    &= \sum_{k=1}^{K} \pi_k \,\mathcal{N}(\nu;\mu_k,\sigma_k^{2}), \\[6pt]
\text{where }\;&
\sum_{k=1}^{K} \pi_k = 1,\quad
\pi_k \ge 0,\quad
\text{and}\quad
\hat{\nu} = \sum_{k=1}^{K} \pi_k \mu_k .
\end{aligned}
\label{eq:UQviaGMM}
\end{equation}

\subsection{Experimental Data Generation}
\label{methods:experimental}
\begin{figure*}[t]
\centering
\includegraphics[scale=1.0]{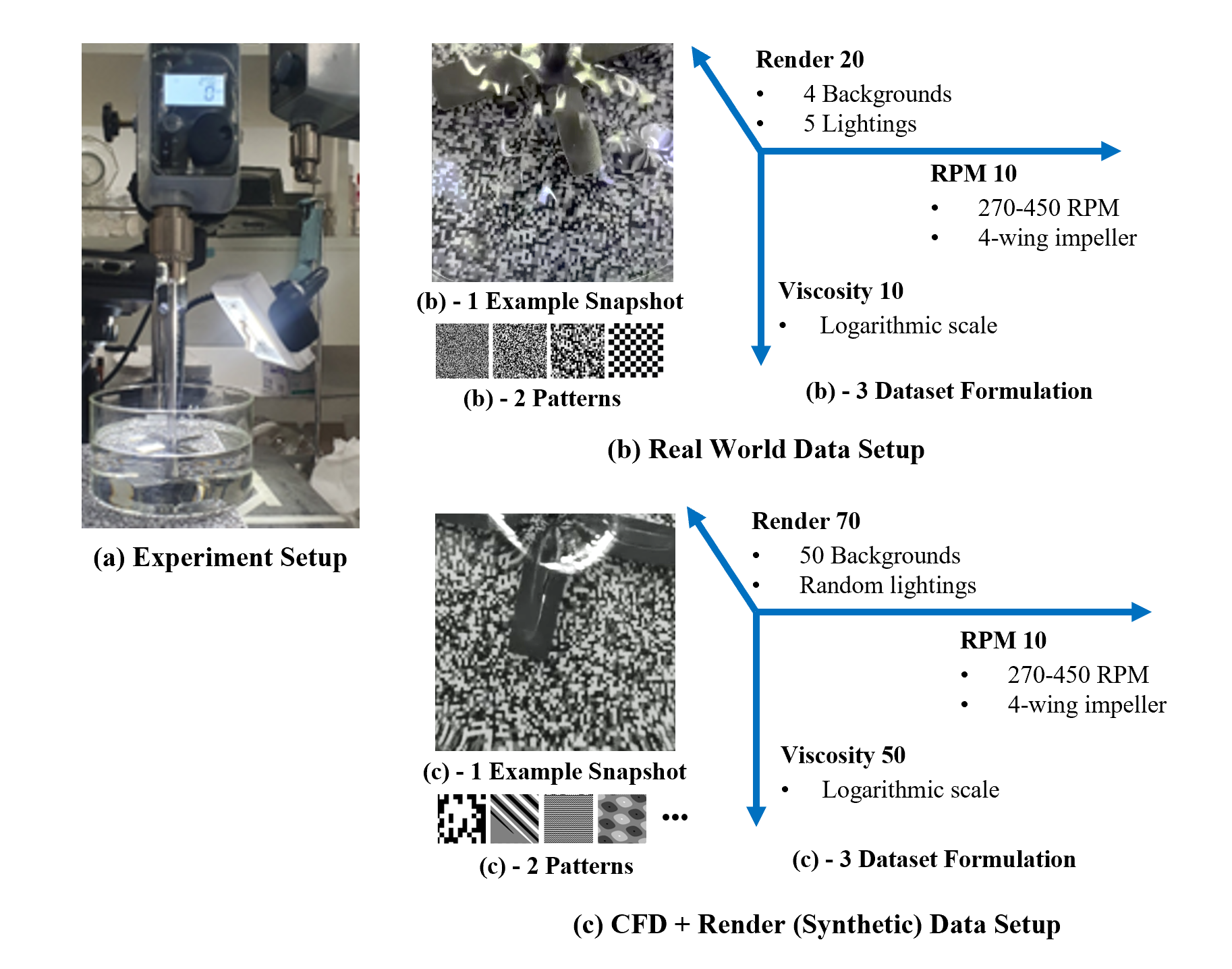}
\caption{Experiment setup and Dataset Diversity Structure.}
\label{fig:dataformulation}
\end{figure*}

A total of 2,000 mixing-experiment videos were generated to train the model, as shown in Fig.~\ref{fig:dataformulation}(a) and Fig.~\ref{fig:dataformulation}(b). A set of 10 glycerin–water mixtures was prepared, covering kinematic viscosity values from $8.96\times10^{-6}$ to $1.30\times10^{-4}\,\mathrm{m^{2}\,s^{-1}}$ at 25 °C~\citep{Simha1949}, logarithmically distributed across this range. Specific values are mentioned in \ref{appendix:A}. The mixtures were stored in a transparent cylindrical container to provide a clear view of the container bottom. A four-blade stainless-steel propeller (diameter: 4 cm) was placed at the center of the container and immersed to a constant depth of 40 mm. The impeller was rotated at 10 equally spaced angular velocities between 270 and 450 RPM, driven by a direct current motor capable of precise speed control. Each trial consisted of steady-state vortex formation under constant rotation, followed by a sudden stop to induce transient decay.

Now four background patterns are placed beneath the container: three white-noise patterns at increasing spatial scales (each twice the scale of the previous) and one checkerboard pattern were utilized. These patterns are referred to as noise-scale-1, noise-scale-2, noise-scale-3, and checkerboard. The resulting vortex scene was recorded using a high-speed camera (iPhone 16) mounted vertically above the container and illuminated under five controlled lighting conditions combining ambient, point, and area light sources. All processed videos were standardized to a 5 s duration, 10 fps frame rate, and 224 $\times$ 224 RGB resolution.

\subsection{Synthetic Data Generation}
\label{methods:synthetic}
Although experimental data provide the most accurate and realistic representation of the mixing process, acquiring such data is time-consuming and inherently limited in diversity, which restricts the generalizability of the trained model. To complement this limitation, we generated additional 35,000 synthetic videos using the divergence-free smoothed particle hydrodynamics (DFSPH) method~\citep{sph, Bender2022}, as shown in Fig.~\ref{fig:dataformulation}(c). Particle simulation and surface generation were performed using the open-source SPlisHSPlasH and SplashSurf libraries developed by the Interactive Computer Graphics team~\citep{sph, Loeschner2023}. The resulting scenes were rendered with Blender’s Cycles engine to achieve a high level of visual realism consistent with the physical setup.

The synthetic environment, including the fluid container, camera viewpoint, video duration, and impeller operation sequence, was designed to closely replicate the real experimental conditions. In the synthetic setting, the viscosity range was kept identical to that of the physical experiments but sampled more finely into 50 logarithmically spaced levels to increase the density of training examples. Background patterns were generated procedurally with 50 random scales and structures, and lighting conditions were randomly varied to enhance visual diversity. This approach allowed us to efficiently create a large, diverse, and physically plausible dataset suitable for training robust models.

\subsection{Training Scheme}
\label{methods:training}

\subsubsection{Curriculum Learning}
\label{methods:training:learning}
We adopt a curriculum learning strategy for stable model training. We first pretrain the encoder on a synthetic dataset for classification to provide inductive bias on when and where to focus on. We then fine-tune the model on 2,000 real fluid videos with the dataset randomly split into a 50\% training set and a 50\% validation set. Standard cross-entropy loss is implemented for training. The baseline model is trained and validated on all real world videos, while four additional models are trained and validated separately on each subset of 500 videos containing only one background pattern.

Next, we train the uncertainty-aware regression layer on top of the pre-trained classification weights without freezing the encoder, allowing the model to more accurately capture continuous viscosity outputs. The regression is performed in the log-viscosity space with $z$-normalization during training for numerical stability, a standard approach in multi-scale regression tasks~\citep{logviscosity}. For training we use the negative log-likelihood (NLL) loss, to model uncertainty.

\begin{equation}
\mathcal{L}_{\text{NLL}}
=
-\frac{1}{N}
\sum_{i=1}^{N}
\log\left(
\sum_{k=1}^{K}
\pi_{ik}\,
\mathcal{N}\!\left(y_i \mid \mu_{ik}, \sigma_{ik}^2\right)
\right),
\end{equation}

Finally, we apply post-hoc calibration by scaling all predicted standard deviations with a single global factor $s$, chosen so that the empirical coverage matches the nominal 50\%, 68\%, and 95\% confidence intervals~\citep{calibration, quantile_caliabration}. Let $D \in \mathcal{X}$ denote the input video data with impeller rotation speed, 
and let $Y \in \mathcal{Y}=\mathbb{R}$ be the ground–truth target. 
The Vivit-GMM model outputs a conditional distribution $p(y \mid D) \in \mathcal{S}$. 
Furthermore, let 
$\hat{F}_{Y \mid D} : \mathcal{Y} \rightarrow (0,1)$ 
denote the corresponding conditional cumulative distribution function (CDF), 
and let 
$\hat{F}^{-1}_{Y \mid D} : (0,1) \rightarrow \mathcal{Y}$ 
denote the associated inverse quantile function.

We then seek a numerical solution to the calibration condition, 

\begin{equation}
\mathbb{P}\!\left(
  Y \le \hat{F}^{-1}_{Y\mid D,s}(\tau \mid D)
\right)
= \tau,
\end{equation}

where the scaled predictive standard deviation is given by

\begin{equation}
\tilde{\sigma}_i = s\, \hat{\sigma}_i .
\end{equation}

\subsubsection{Metrics for Analysis}
\label{methods:training:metrics}
First, to interpret how the model understands viscosity, we use dimensionless numbers that summarize the dominant fluid motion into a single scalar. These values are then used to analysis model performance with attention maps and classification accuracy. We visualize and analyze both temporal and spatial attention maps from the final ViViT layer.

In the mixing setup, the free-surface dynamics is governed by inertial forces from the impeller, alongside with viscous and capillary forces inherent to the fluid. The relevant dimensionless ratios are the Reynolds (\textit{Re}), Capillary (\textit{Ca}), and Weber number (\textit{We}). The Reynolds number characterizes the balance between inertia and viscous dissipation, and thus distinguishes laminar, transitional, and turbulent surface motions. The Capillary number captures the competition between viscous stresses and surface-tension restoring forces, governing interface stretching and meniscus deformation. The Weber number quantifies the ratio of inertial to capillary forces and indicates the fluid’s propensity toward surface breakup, wave steepening, and free-surface instabilities under strong impeller-driven acceleration. Note that the Weber number does not have dependency on viscosity.

\begin{equation}
\begin{aligned}
\mathrm{\textit{Ca}} &= \frac{\rho\, \nu\, (\omega R)}{\sigma}, 
\qquad
\mathrm{\textit{Re}} = \frac{(\omega R)\, L}{\nu}, 
\qquad
\mathrm{\textit{We}} = \frac{\rho\, (\omega R)^{2}\, L}{\sigma},\\[10pt]
\text{where }\;&
\nu \text{ is the kinematic viscosity (m$^{2}$/s)},\\[4pt]
&\rho \text{ is the density (kg/m$^{3}$)},\\[4pt]
&\omega \text{ is the angular speed (rad/s)},\\[4pt]
&R \text{ is the characteristic radius (m)},\\[4pt]
&\sigma \text{ is the surface tension (N/m)},\\[4pt]
&L \text{ is the characteristic length scale (m)}.
\end{aligned}
\end{equation}

Second, to evaluate the validity of uncertainty estimation before calibration, we use three primary metrics: mean absolute error (MAE) for prediction accuracy, area under the sparsification error curve (AUSE) for error-uncertainty alignment and calibration error (CE) for coverage fit. In detail, MAE is computed in the $\log \nu~(\mathrm{m}^2\,\mathrm{s}^{-1})$ space, and used as the error value in the $y$-axis in sparsification plots. In calculating AUSE, we normalize the total error so that the maximum value becomes $1$. Moreover, because AUSE is inherently relative, we compare it against the AUSE obtained from randomized uncertainty assignments ~\citep{AUSE2, AUSE1}. Then we conduct post-hoc calibration and perform coverage assessment using CE. Using the definition of CDF as in Equation 5 and nominal quantile level $\{\tau_j\}_{j=1}^{M}$, CE is written as \citep{UQmetrics}

\begin{equation}
\mathrm{CE} = \sum_{j=1}^{M} \,\left( \tau_j - \hat{\tau}_j \right)^2.
\label{eq:cal}
\end{equation}
\section{Results}
\label{results}
\subsection{Classification Model}
\label{results:classification}

\subsubsection{Performance}

\begin{figure*}[t]
\centering
\includegraphics[scale =0.8]{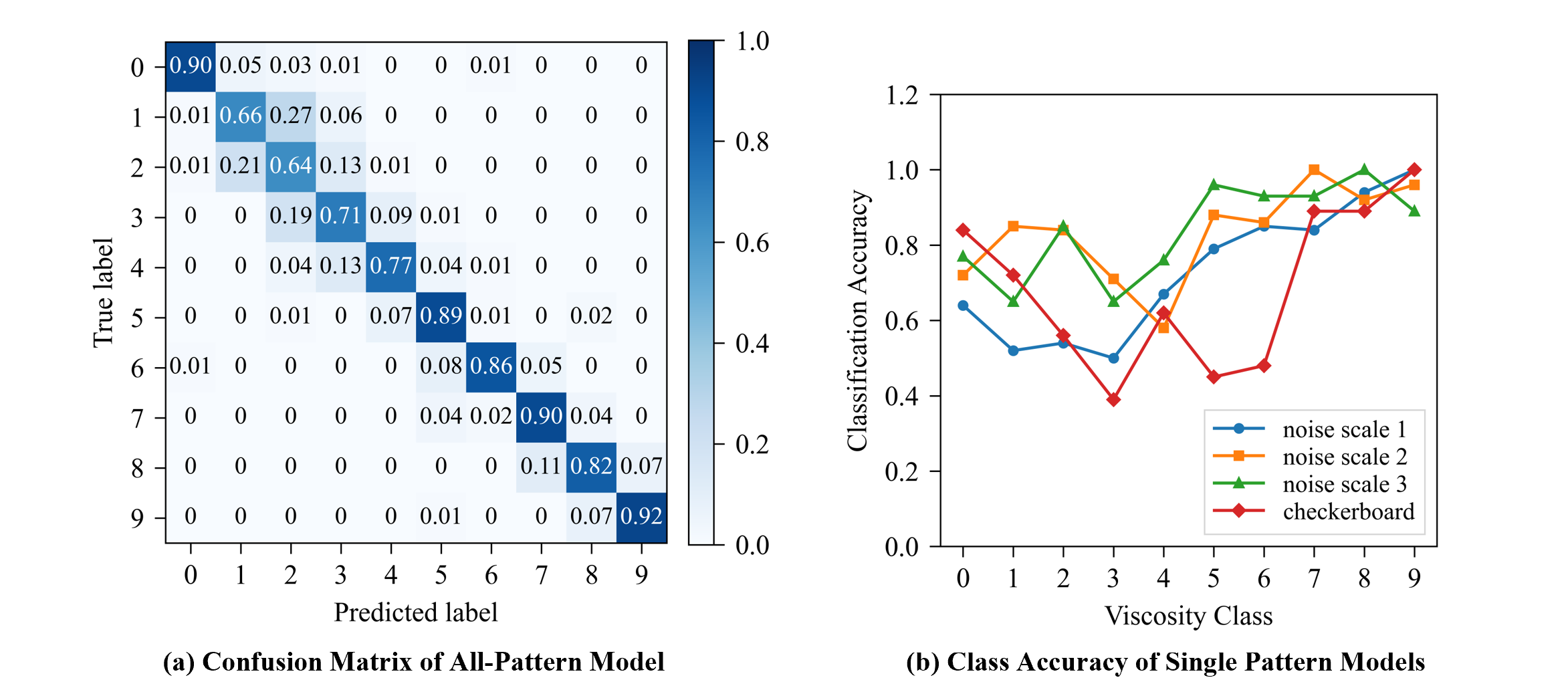}
\caption{(a) Confusion matrix of the classification model trained with all 4 patterns, (b) Class accuracy plotted for each single pattern models.}
\label{fig:confusion_matrix}
\end{figure*}

\begin{figure*}[!ht]
\centering
\includegraphics[scale =0.8]{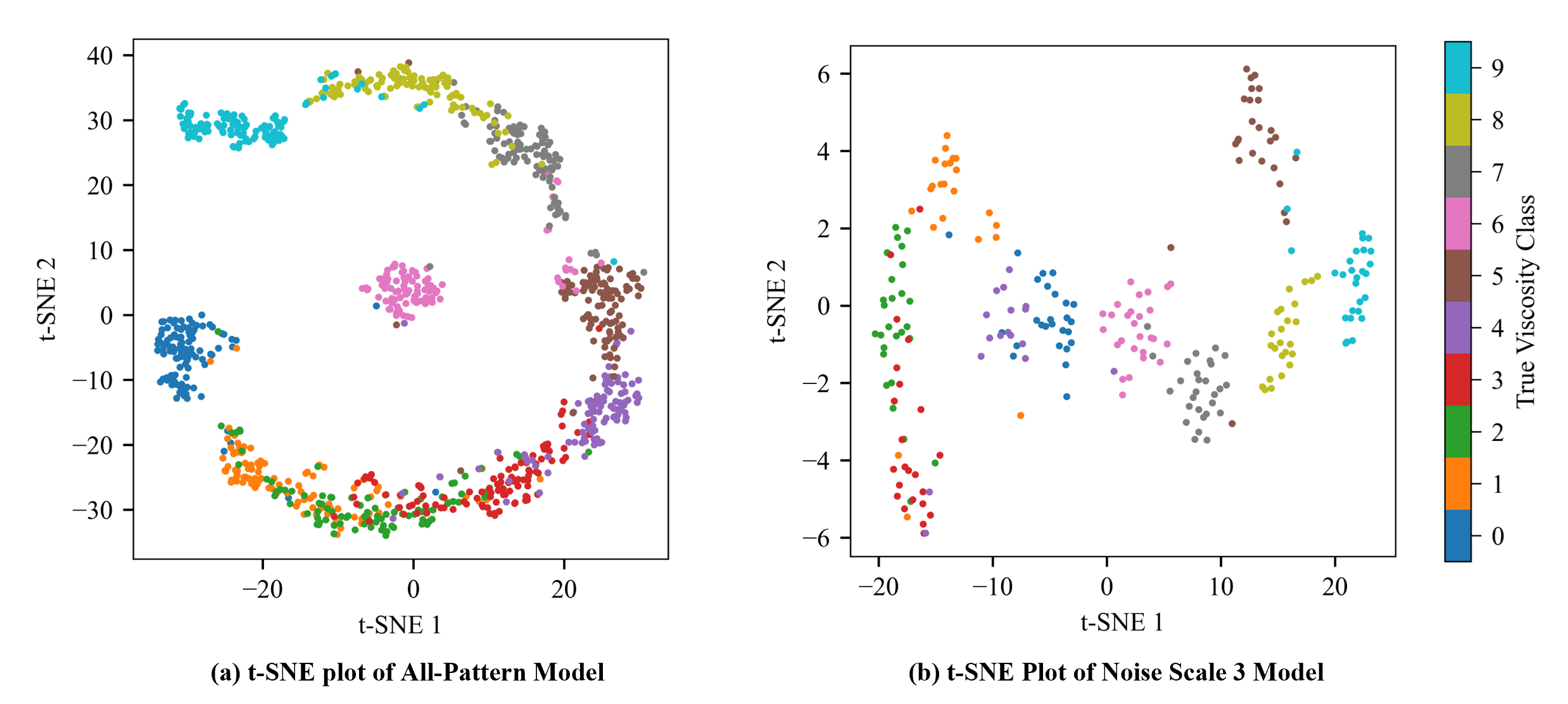}
\caption{t-SNE plots of (a) the all-pattern model and (b) the noise-scale-3 model, which achieved the highest performance.}
\label{fig:tsne}
\end{figure*}

The performance of the trained classifier models was evaluated as shown in Fig.~\ref{fig:confusion_matrix}. The model trained with 1,000 videos of all patterns, reached 80.7\% accuracy, and the best-performing single-pattern trained model (noise, scale 3) achieved 83.9\% accuracy. Although the result varies across different background patterns, all models consistently show reduced performance in distinguishing classes 1, 2, and 3, which correspond to kinematic viscosities of \(1.36\times10^{-5},\, 1.80\times10^{-5},\, 2.38\times10^{-5}\,\mathrm{m}^2 \mathrm{s}^{-1}\). It is also evident in the t-SNE plots in Fig.~\ref{fig:tsne} that the encoded features for viscosity classes 1, 2, and 3 are not clearly separable from each other, whereas the data points for the other viscosities form class-specific clusters.

\subsubsection{Physical Interpretation of Errors}

\begin{figure*}[t]
\centering
\includegraphics[scale =0.9]{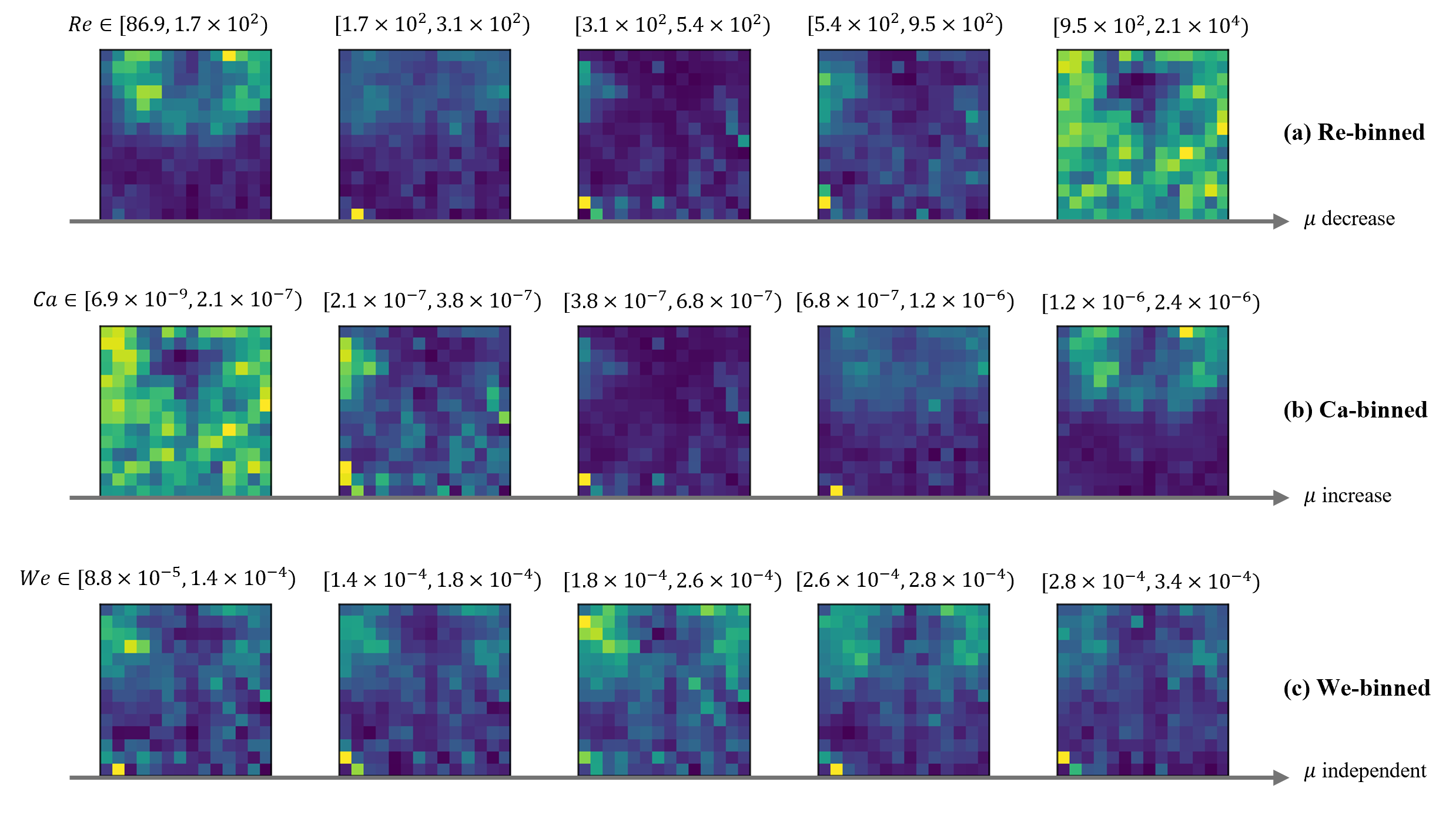}
\caption{Spatial attention graphs of the encoder’s final layer, binned into five groups by (a) Reynolds number, (b) Capillary number, and (c) Weber number.}
\label{fig:attnmap_align}
\end{figure*}

To understand why errors increase in certain viscosity ranges, we first examine how the physical conditions present in the dataset relate to the visual information perceived by the model. Although the model is trained to predict viscosity, it may implicitly encode additional fluid properties including surface tension and density. Therefore, we analyze the model’s response to changes in the three dimensionless number: \textit{Re}, \textit{Ca}, and \textit{We}. Each video is assigned its corresponding values of these numbers, and those with similar magnitudes are grouped to visualize the average attention pattern within each region in Fig.~\ref{fig:attnmap_align}. Vortices with small  Reynolds number only has smooth surface deformations near the center of the impeller, causing the model to attend in those regions. On the other hand, when the Reynolds number is large, turbulence increases, generating more eddy flows in the outer region, which shifts the model’s attention outwards. The Capillary number, which scales directly with viscosity (in contrast to the inverse dependency of Reynolds number), consequently exhibits a visual trend opposite to that of the Reynolds number. In contrast, the Weber number, being independent of viscosity, shows no clear trend in the model’s attention. From these observations, we conclude that viscosity is the primary determinant of the visual features perceived by the model.

This is also an intutive result, where physically higher viscosity accelerates energy dissipation, dampening high-frequency turbulent fluctuations and resulting in smoother optical distortions. Conversely, lower viscosity sustains smaller eddies, generating chaotic refractive patterns.

\begin{figure}[t]
\centering
\includegraphics[scale=1.0]{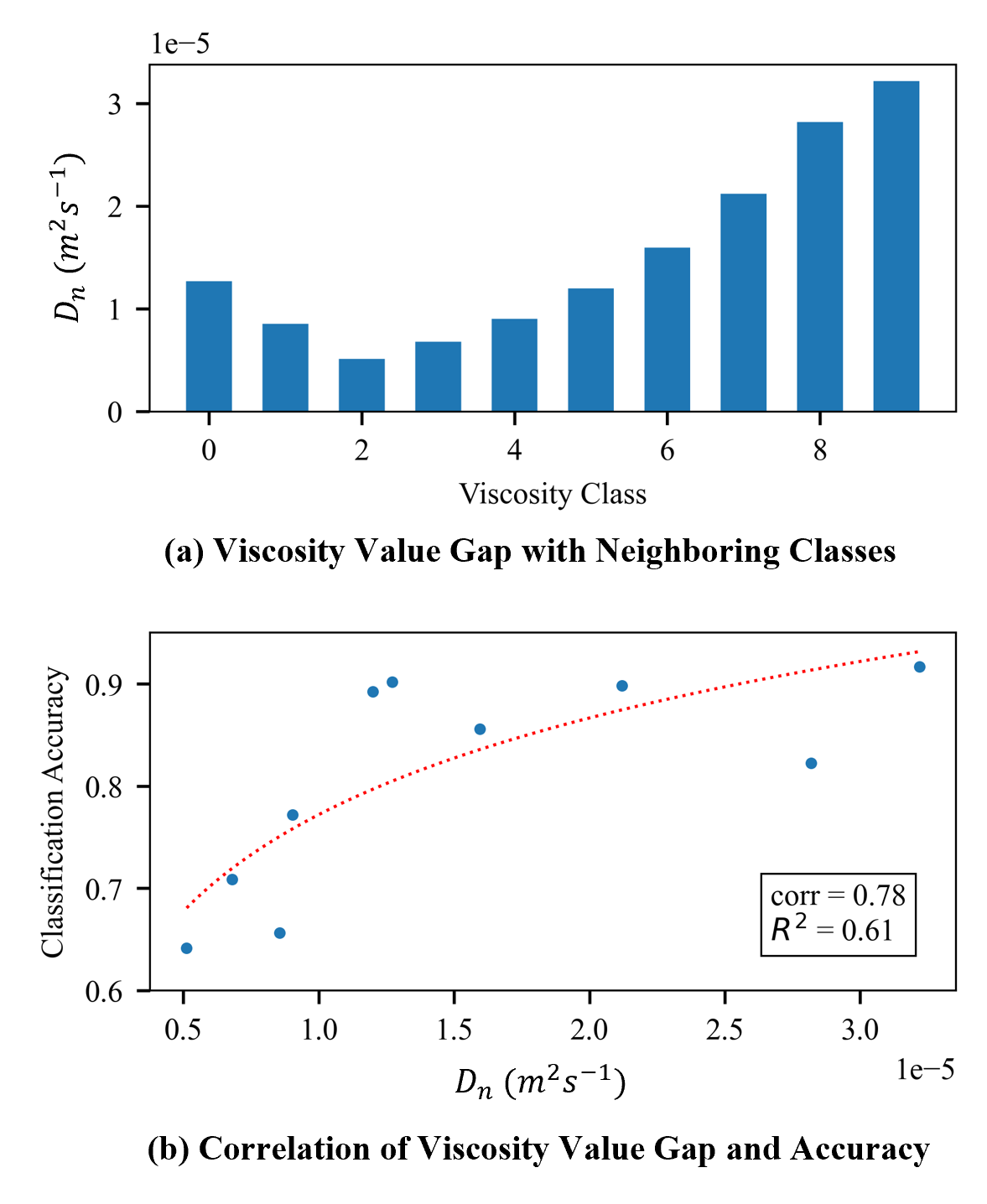}
\caption{(a) Average of absolute viscosity value difference $D_n$ with adjacent classes, (b) Logarithmic regression between viscosity value gap $D_n$ and classification accuracy of each class.}
\label{fig:proximity}
\end{figure}

This information implies that the model would misclassify videos of fluids that have relatively close viscosity values. We calculated how closely the viscosity value of class $n$ lies relative to their neighboring classes as $D_n$. The exact mathematical definition of $D_n$ is given in Eq.~\eqref{eq:proximity}. Figure~\ref{fig:proximity}(a) shows that viscosity values near class 2 are densely clustered. Also, this degree of proximity exhibits a high positive correlation with classification accuracy as shown in Fig.~\ref{fig:proximity}(b). However, the relationship is logarithmic rather than linear, because separation of class viscosity values beyond a certain level yields little additional improvement in accuracy.

\begin{equation}
\begin{aligned}
D_n =
\begin{cases}
\lvert \nu_{n} - \nu_{n+1} \rvert, & n = 0, \\[6pt]
\displaystyle \frac{\lvert 2\nu_{n} - \nu_{n-1} - \nu_{n+1} \rvert}{2}, 
& n = 1, 2, \dots, 8, \\[8pt]
\lvert \nu_{n} - \nu_{n-1} \rvert, & n = 9 .
\end{cases}
\end{aligned}
\label{eq:proximity}
\end{equation}

This interpretation is physically well supported: adjacent viscosity values result in nearly identical viscous momentum transfer, producing interface geometries that generate optically indistinguishable distortion patterns under fixed impeller rotation speed. This significanty increases the influence of intrinsic noise, such as camera sensor artifacts and surface bubbles, often modeled as Gaussian distributions. Thus, we assume that the main source of prediction error arises from aleatoric heteroscedastic uncertainty.

\subsubsection{Multi-Pattern for Data Enrichment}
\begin{figure}[t]
\centering
\includegraphics[scale = 1.0, keepaspectratio]{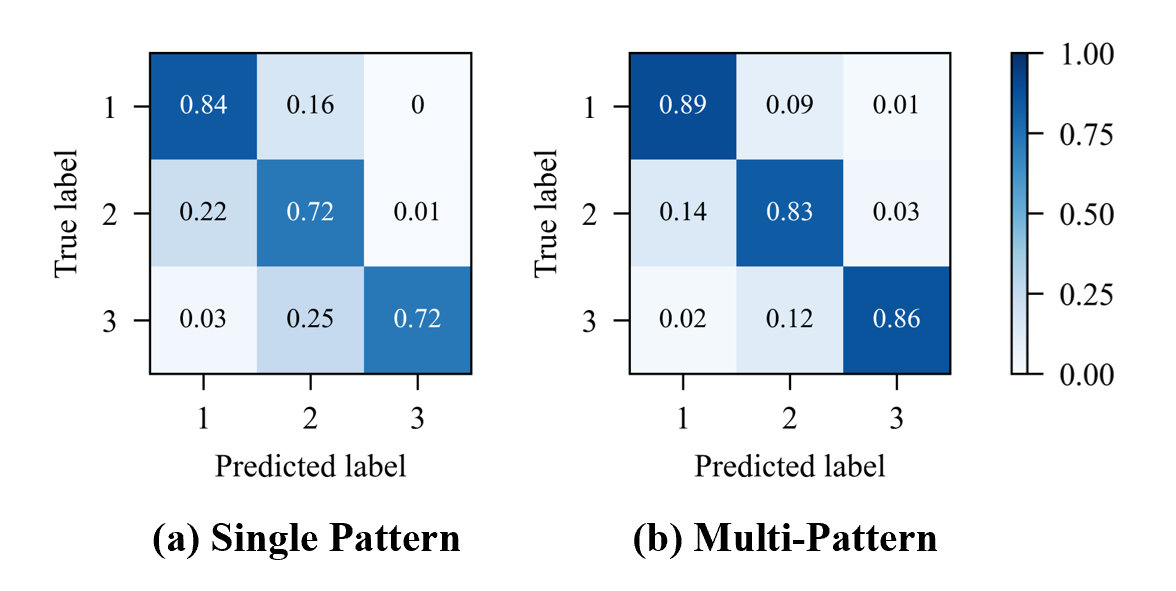}
\caption{Accuracy improvement after multi-pattern placement. The models are trained only on underperforming class 1, 2 and 3. (a) Mono pattern real world finetuned, (b) multi-pattern real world finetuned.}
\label{fig:multipattern}
\end{figure}

Now we propose a multi-pattern approach that enhances prediction accuracy in the previously underperforming region. As viscosity is extracted from how the optical mapping function $f$ (Eq.~\eqref{eq:problem_formulation}) distorts the input signal, providing richer input patterns would directly improve performance. Thus, we place two different background patterns side by side in parallel. The combinations of the patterns are presented in \ref{appendix:B}. Although the precise relationship between pattern geometry and classification accuracy has not yet been fully characterized, it is evident that the geometric features of the background pattern significantly influence the visualization of the free surface \citep{Raffel2015, Raffel2023}. Because the video scene is symmetric about the vertical centerline, we construct mixed-pattern inputs by placing two different patterns side by side and retrain a compact model targeting the three underperforming classes. The results are demonstrated in Fig.~\ref{fig:multipattern}, where multi-pattern approach yielded an accuracy improvement of 13.3\%p compared to the mono-pattern baseline. 

\subsection{Regression and Uncertainty Quantification}
\label{results:uq}
The results of uncertainty quantification are summarized in Table~\ref{tab:uq_metrics}. Notably, the model attains an MAE of 0.113 in $\log\mathrm{m}^2 \mathrm{s}^{-1}$ units, which is below the log-viscosity spacing of 0.123 from linearly spaced target values, indicating accurate regression performance in log space. For error–uncertainty alignment, the AUSE metric shows a 61.3\% improvement when uncertainty modeled, with modeled AUSE much smaller than random AUSE, indicating substantially stronger correspondence between prediction error and modelled uncertainty. 

\begin{table}[!ht]
\centering
\begin{tabular}{cccc}
\hline
\textbf{MAE} & \textbf{AUSE modeled/random} & \textbf{CE before} & \textbf{CE after} \\
\hline
  0.113 & 0.303/0.785 & 0.1154 & 0.0104 \\
\hline
\end{tabular}
\caption{Uncertainty quantification results, MAE is calculated in $\log\mathrm{m}^2 \mathrm{s}^{-1}$.}
\label{tab:uq_metrics}
\end{table}

\begin{figure}[ht]
\centering
\includegraphics[scale =0.9]{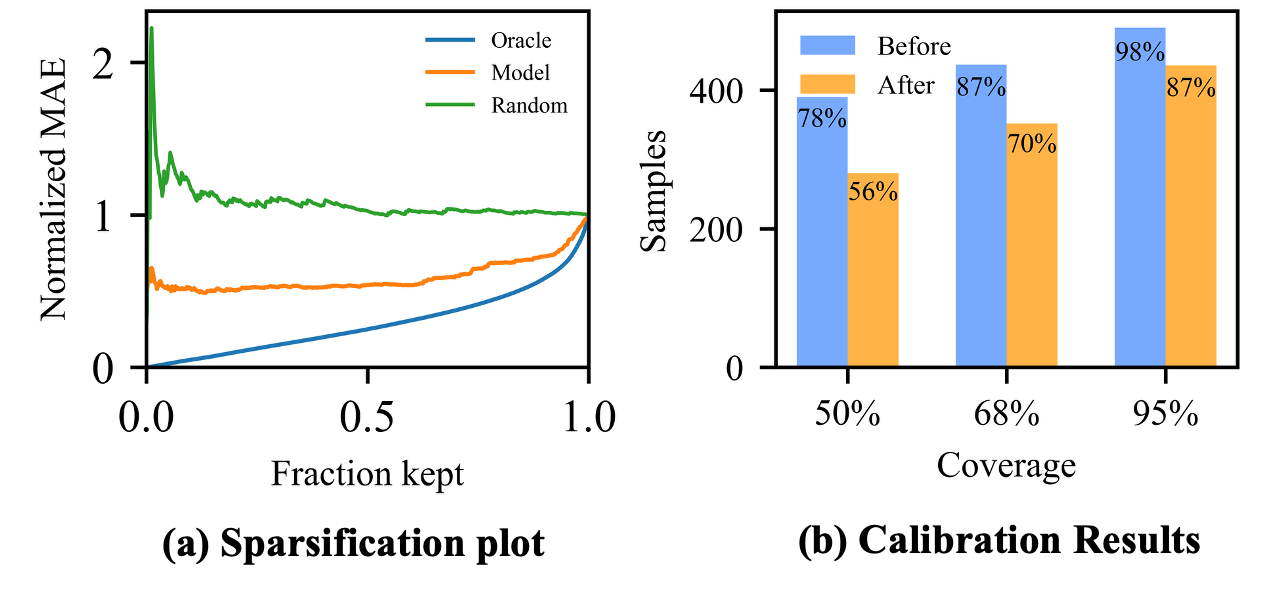}
\caption{(a) Sparsification plot, the modeled curve shows uncertainty aligned with errors values. (b) Coverage check on intervals of 50, 68, and 95\%, after calibration results fit with the expect intervals.}
\label{fig:UQ}
\end{figure}

The sparsification plots with normalized MAE are shown in
Fig.~\ref{fig:UQ}(a). In the plots, the modeled sparsification curve decreases with moving from right to left, while the random baseline remains significantly higher, indicating that the predicted uncertainties successfully identify high-error samples. 

The calibration results are presented in Fig.~\ref{fig:UQ}(b), where the empirical coverage matches the expected coverage. CE values drop after calibration to 0.0104, demonstrating that the predictive distribution is well modeled and calibrated. The optimal scaling factor was $\alpha=0.504$.

\subsection{Data Efficiency}
\label{results:dataefficiency}
\begin{figure*}[t]
\centering
\includegraphics[scale = 0.7]{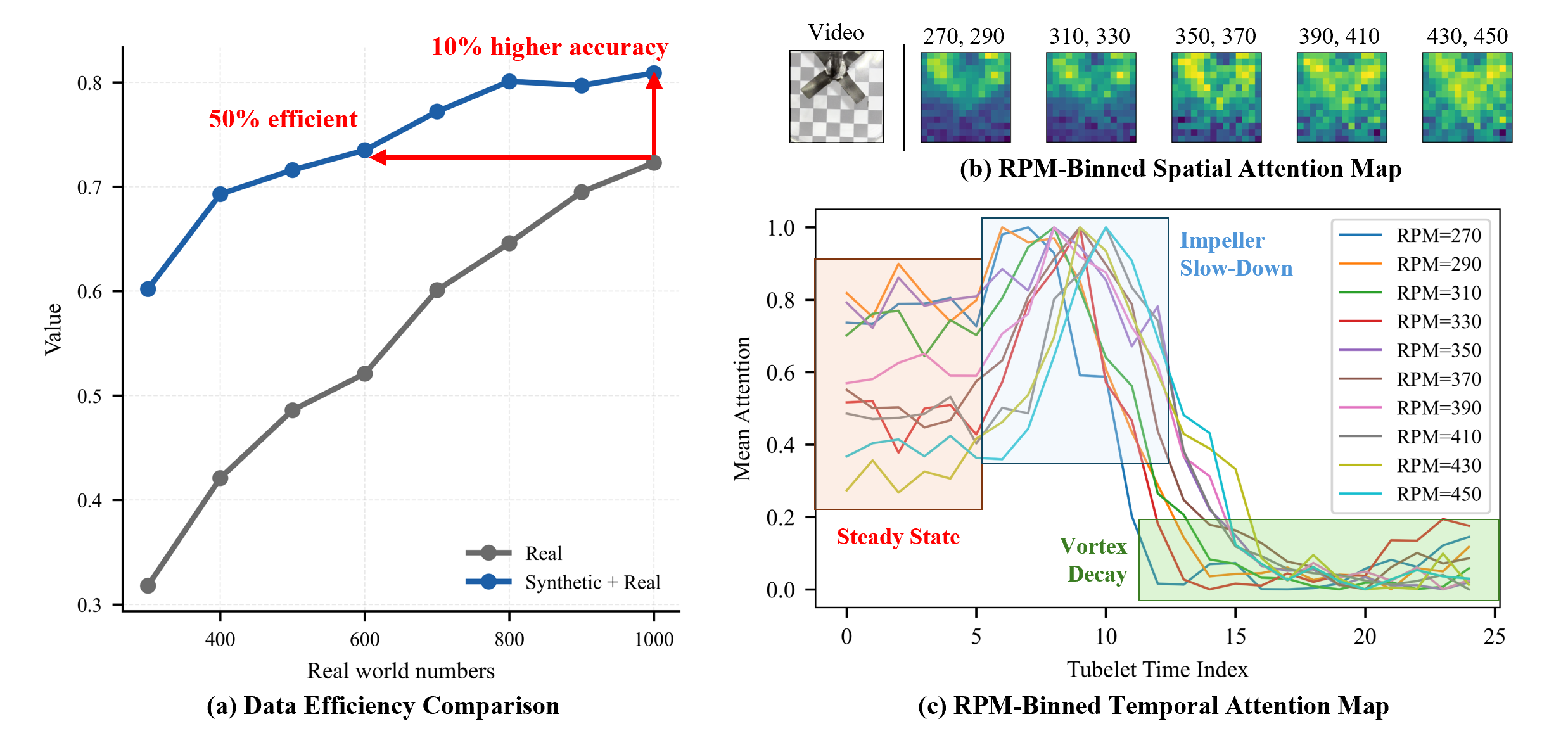}
\caption{(a) Classification accuracy based on the number of real world datasets used. Synthetic data pre-trained models shows better accuracy and data efficiency, (b) Spatial and (c) temporal attention maps averaged across spatial and temporal dimensions, conditioned on impeller rotation speed, $\omega$ (RPM). The model has inductive bias on where and when to attend.}
\label{fig:dataefficiency}
\end{figure*}

Finally, we demonstrate the practical benefit of incorporating the simulation dataset through classification accuracy results. As shown in Fig. \ref{fig:dataefficiency}(a), the model trained on both synthetic and real data achieves a 10\%p higher maximum accuracy and reaches the same accuracy with 50\% less real data points compared to the model trained only on real data.

Synthetic datasets inherently exhibit lower fidelity; for instance, the DFSPH method relies on modeling assumptions, and volumetric point-cloud representations are mathematically smoothed into mesh surfaces, causing the loss of fine details. Nonetheless, the performance gains arise because the synthetic data provide macroscopic prior knowledge of the flow dynamics. The spatial and temporal attention maps in Fig.~\ref{fig:dataefficiency}(b), generated after training with synthetic data and evaluated on real world validation sets, further show that synthetic data guides the model to focus on impeller-driven regions in turbulent flows and to capture decay dynamics associated with changes in impeller speed, particularly where decay begins near the 8th tubelet.

\section{Final Remarks}
\label{FinalRemarks}

In this work, we developed a sensor capable of inferring bulk fluid viscosity by decoding the optical distortions induced by the mixing vortex surface acting on a background pattern. A key insight is that mixing-driven free-surface deformation naturally encodes viscosity by refractively distorting a fixed background pattern, enabling stand-off, \textit{in-situ} viscometry.

Several limitations and opportunities for further improvement still remain. The current model is restricted to transparent Newtonian fluids, as free-surface visualization depends on optical clarity and interfacial properties; thus, fine-tuning will be required for different applications and setups. The method also requires intermittent pauses in mixing to observe viscous dissipation and depends on knowledge of the impeller’s rotational speed. Although this remains far more convenient than extracting fluid samples, it is not yet fully in-line. Future work should therefore incorporate more diverse mixing sequences to address this limitation. In the single-pattern models, we observed that performance varies with background pattern selection, although no clear trend was identified. Subsequent studies may parameterize pattern characteristics to determine which features of the fluid surface are most effectively visualized by each pattern and how this affects model performance. Furthermore, identifying which pair of patterns yields the best results within the multipattern framework presents another promising research direction.

\section*{Acknowledgments}
The authors express special thanks to Joonhee Oh for discussions in machine learning topics.
This work was supported by SNU Student-directed Education Undergraduate Research Program through the Faculty of Liberal Education, Seoul National University (2025). It is also supported by the National Research Foundation of Korea (NRF) grant funded by the Korea government (Ministry of Science and ICT, MSIT) (grant numbers RS-2021-NR057526, RS-2023-NR077027).

\section*{Competing Interest Statement}
The authors declare no competing interest.

\section*{Data Availability}
All data supporting the results of this study will be made available upon request.

\appendix
\section{Fluid Properties of Experimental Dataset}
\label{appendix:A}

Table.~\ref{tab:fluid_properties} shows the properties of the water-glycerin solution used for the generation of the experiment dataset. Kinematic viscosity is distributed logarithmically from $8.96\times10^{-6}$ to $1.30\times10^{-4}\,\mathrm{m^{2}\,s^{-1}}$ across classes 1–9, while class 0 corresponds to pure water and serves as a high-turbulence reference. 

\renewcommand{\arraystretch}{1.4}
\begin{table*}[!ht]
\centering
\begin{tabular}{c c c c c}
\hline
\shortstack{\rule{0pt}{3ex}Viscosity\\class} &
\shortstack{\rule{0pt}{3ex}Glycerin\\weight fraction} &
\shortstack{\rule{0pt}{3ex}Density\\(kg·m$^{-3}$)} &
\shortstack{\rule{0pt}{3ex}Kinematic viscosity\\($10^{-6}$ m$^{2}$·s$^{-1}$)} &
\shortstack{\rule{0pt}{3ex}Surface tension\\(N·m$^{-1}$)} \\
\hline
0 & 0.000 & 996.890 & 0.89552 & 0.07280 \\
1 & 0.684 & 1173.76 & 1.36066 & 0.06637 \\
2 & 0.720 & 1183.44 & 1.80070 & 0.06603 \\
3 & 0.753 & 1192.29 & 2.38489 & 0.06572 \\
4 & 0.783 & 1200.40 & 3.16070 & 0.06544 \\
5 & 0.810 & 1207.86 & 41.9134 & 0.06518 \\
6 & 0.836 & 1214.75 & 55.6089 & 0.06494 \\
7 & 0.859 & 1221.12 & 73.8131 & 0.06472 \\
8 & 0.881 & 1227.03 & 98.0165 & 0.06452 \\
9 & 0.902 & 1232.51 & 130.203 & 0.06432 \\
\hline
\end{tabular}
\caption{Physical properties of glycerin water mixtures used in the experimental dataset.}
\label{tab:fluid_properties}
\end{table*}
\renewcommand{\arraystretch}{1.0}

\section{Multi-Pattern Dataset Examples}
\label{appendix:B}

\begin{figure}[!ht]
\centering
\includegraphics[scale = 1.0]{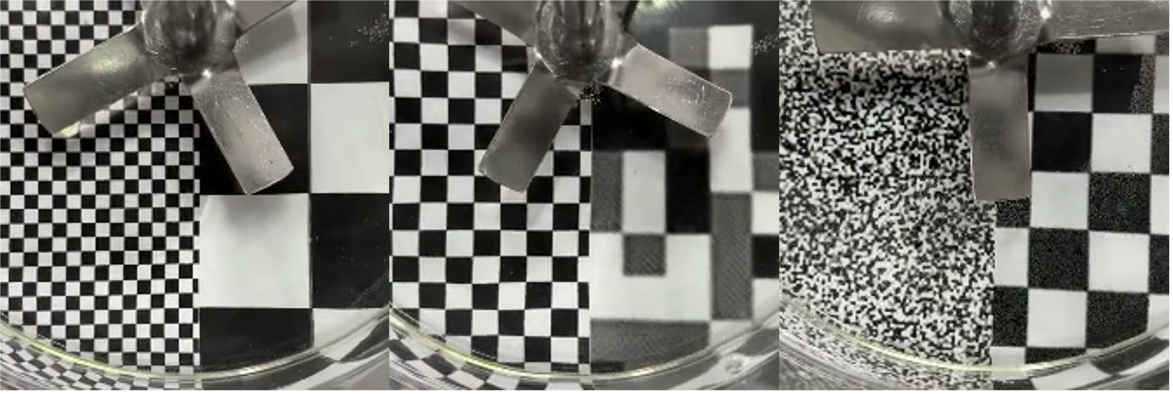}
\caption{Each are examples of the multi pattern datasets implemented for data enrichment.}
\label{fig:multipatternexamples}
\end{figure}

The multi-pattern datasets were created by combining the patterns shown in Fig. \ref{fig:multipatternexamples}. Although the design space of pattern combinations was not exhaustively explored, we intentionally paired a coarse pattern with a finer one and positioned the coarse pattern on the right.

\bibliographystyle{elsarticle-harv}
\bibliography{elsarticle/references}
\end{document}